\pdfoutput=1

\documentclass[11pt]{article}
\usepackage{CJK}
\usepackage{ACL2023}
\usepackage{subcaption}
\usepackage{times}
\usepackage{latexsym}
\usepackage{makecell}
\usepackage[T1]{fontenc}

\usepackage[utf8]{inputenc}
\usepackage{graphicx}
\usepackage{microtype}

\usepackage{inconsolata}

\usepackage{makecell}
\usepackage{multirow}
\usepackage{amsmath}
\usepackage{amssymb}
\usepackage{enumitem}
\usepackage{color}
\usepackage{graphicx}
\usepackage{comment}

\usepackage{geometry}
\usepackage{algorithm2e}

\setenumerate[1]{itemsep=0pt,partopsep=0pt,parsep=\parskip,topsep=3pt}
\setitemize[1]{itemsep=0pt,partopsep=0pt,parsep=\parskip,topsep=3pt}

\title{Unified Text Structuralization with Instruction-tuned Language Models}

\author{Xuanfan Ni$^1$, Piji Li$^1$, and Huayang Li$^2$\\
  $^1$Nanjing University of Aeronautics and Astronautics\\
  $^2$Nara Institute of Science and Technology \\
  $^1$\texttt{\{xuanfanni, pjli\}@nuaa.edu.cn}, $^2$\texttt{li.huayang.lh6@is.naist.jp}}

\begin{document}
\maketitle
\begin{abstract}
Text structuralization is one of the important fields of natural language processing (NLP) consists of information extraction (IE) and structure formalization. However, current studies of text structuralization suffer from a shortage of manually annotated high-quality datasets from different domains and languages, which require specialized professional knowledge. In addition, most IE methods are designed for a specific type of structured data, e.g., entities, relations, and events, making them hard to generalize to others. In this work, we propose a simple and efficient approach to instruct large language model (LLM) to extract a variety of structures from texts. More concretely, we add a prefix and a suffix instruction to indicate the desired IE task and structure type, respectively, before feeding the text into a LLM. Experiments on two LLMs show that this approach can enable language models to perform comparable with other state-of-the-art methods on datasets of a variety of languages and knowledge, and can generalize to other IE sub-tasks via changing the content of instruction.
Another benefit of our approach is that it can help researchers to build datasets in low-source and domain-specific scenarios, e.g.,  fields in finance and law, with low cost.
\end{abstract}

\section{Introduction} \label{introduction}
\begin{figure*}[!t]
   \centering
   \includegraphics[width=\linewidth]{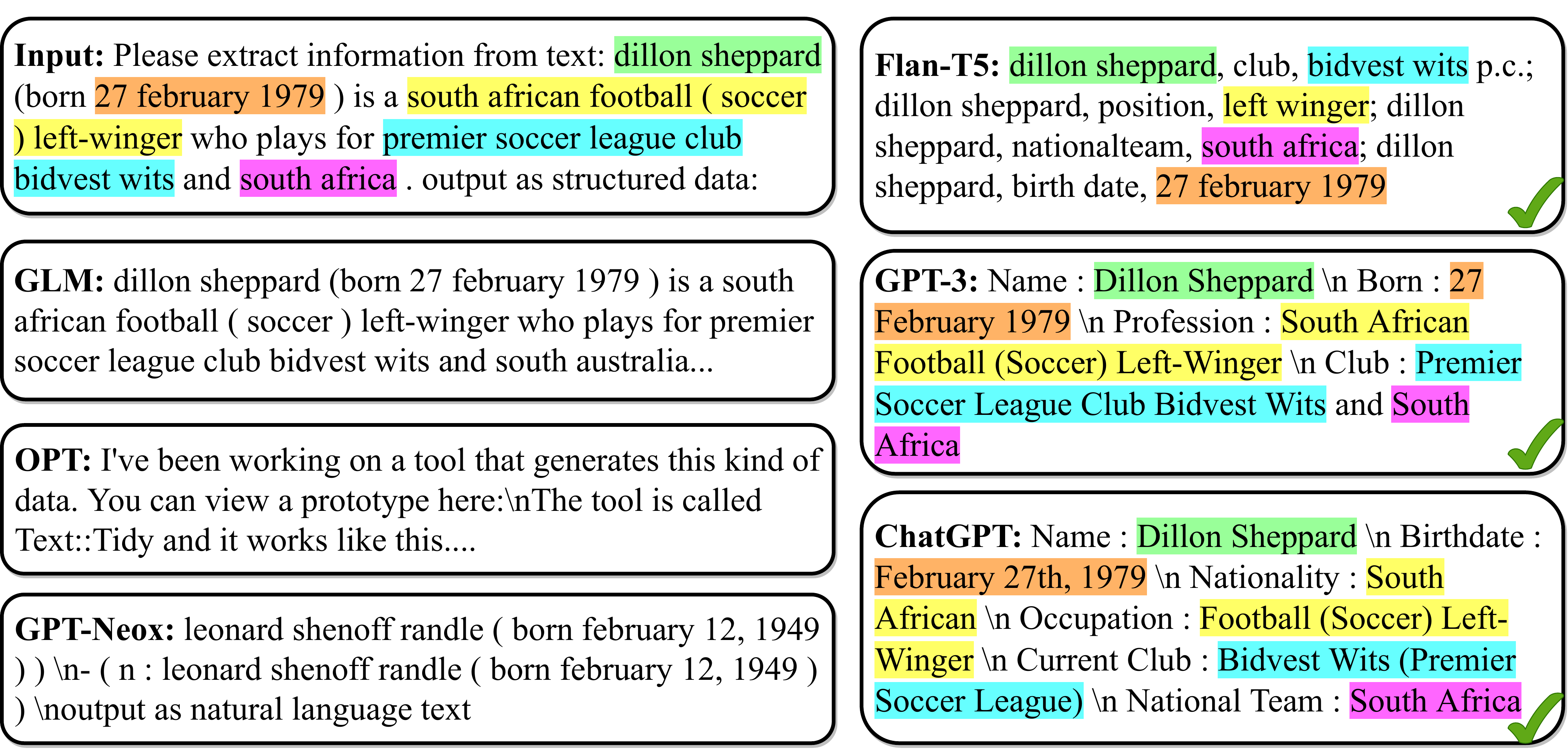}
    \caption{Outputs of different models for the same text with instructions. Text highlighted in the same colour indicates the same key information. A green tick indicates that the output is structured and correct.}
    \label{example}
\end{figure*}

Text structuralization is a task aims to convert unstructured natural language to structured data. These structured results are beneficial for many downstream tasks, such as data mining and text analysis. Text structuralization generally has the following two steps: (1) Information extraction (IE), such as named entity recognition (NER) (\citealp{huang2015bidirectional}; \citealp{ma2016end}; \citealp{devlin2018bert}), relation extraction (RE) (\citealp{zeng2018extracting}; \citealp{luan2019general}; \citealp{zhong-chen-2021-frustratingly}), event extraction (EE) (\citealp{yang2019exploring}; \citealp{wadden2019entity}), etc.; (2) Structure Formalization, which change the format of the information extracted from text to key-value pairs. Table \ref{text example} shows an example of a text converted to structured data.

\begin{table}[!t]
\resizebox{\linewidth}{!}{
\centering
\begin{tabular}{lp{5cm}}
\Xhline{3\arrayrulewidth}
\textbf{Input Text}  & \begin{tabular}{@{}p{5cm}@{}}dillon sheppard ( born 27 february 1979 ) is a south african football ( soccer ) left-winger who plays for premier soccer league club bidvest wits and south africa\end{tabular} \\
\hline
\textbf{Output Result}    & \begin{tabular}{@{}l@{}} Name: dillon sheppard\\ Birth date: 27 february 1979\\ Nationality: south african\\ Occupation: football left-winger\\ Club: bidvest wits\\ National team: south africa \end{tabular}\\
\Xhline{3\arrayrulewidth}
\end{tabular}
}
\caption{An example of a text to structured data, in which we first extract crucial information from text, e.g., name entites, events and relations, and generate the appropriate key to form a key-value pair for each extracted content.}
\label{text example}
\end{table}

The studies of text structuralization and information extraction have been developed for decades (\citealp{wu2020keyword}; \citealp{wang2019text};), and current techniques can achieve high performance on amount of benchmark datasets and tasks. However, there are still the vast amounts of difficulties and challenges.
First, building high-quality labelled datasets for text structuralization tasks is extremely expensive, because the tasks may cover many different domains and require a lot of professional human effort. For instance, ACE2005 \citep{doddington2004automatic} dataset, which is the most popular dataset among IE studies, only contains 599 documents for training. This may result in insufficient training data and poor model performance. Second, datasets for text structuralization may vary widely in domains and languages. However, most methods in text structuralization may be biased to a specific dataset, which results in a poor generalization ability. An example can well illustrate the above point that Bert-base-NER\footnote{https://huggingface.co/dslim/bert-base-NER}
\citep{DBLP:journals/corr/abs-1810-04805} can achieve an F1-score of 91.3\% on test set of CoNLL-2003 \citep{sang2003introduction}, but only gets F1-score less than 10\% on test set of Gutenberg Time \citep{kim2020time}, an english time NER dataset.



In addition, the existing deep learning based text structuralization models can be divided into different folds, such as pipeline \citep{zhong-chen-2021-frustratingly}, end2end \citep{ma2016end}, joint modeling \citep{lin2020joint}, etc. And most of them require multiple sub-models to be cascaded, like PURE \cite{zhong-chen-2021-frustratingly}, which uses two separate models for entity extraction and relation classification tasks. This is complicated and can lead to errors accumulating and propagating between models. Constructing a general model that can simultaneously extract entities, relations, events, and other crucial information is important but underexplored.

Recently, large language models have achieved amazing success in the field of NLP. Instruction-tuning (\citep{brown2020language}; \citealp{chung2022scaling}), which further improves the generalization ability of large models and their performance under zero-shot condition, comes from the intuition that NLP tasks can be described by natural language instructions, such as \emph{Answer the following yes/no question by reasoning step-by-step } or \emph{Translate the text into English}. Leveraging large language models to cope with NLP tasks can obtain several advantages: (1) Large language models have a large number of parameters and a rich training corpus. Therefore, they can store expertise in different domains and have comprehension and generalisation capabilities to deal with cross-domain tasks. (2) Benefiting from their instruction tuning and fine-tuning training method, large language models can achieve good performance by providing a few examples or even an instruction without training.



Inspired by the advantages of the instruction-tuned model, to address the above challenges and issues, we propose a straightforward instruction-based approach to conduct the problem of zero-shot/few-shot information extraction and text structuralization. Specifically, we guide large language models to perform text structuralization task by adding a prefix instruction and a suffix instruction. Following this idea, we test 6 large language models, namely ChatGPT, GPT-3 (InstructGPT, 175B) \citep{brown2020language}, FLAN-T5-XXL(13B) \citep{chung2022scaling}, GPT-NeoX-20B \citep{gpt-neo}, OPT-30B \citep{zhang2022opt} and Glm-130B \citep{zeng2022glm}. Figure \ref{example} shows the results of these large language models, from which we can see that only the instruction-tuned models, GPT-3, ChatGPT and FLAN-T5, have the ability to recognize instructions and perform corresponding tasks. We conduct text structuralization experiments on four datasets, namely Wikibio, WikitableT, Rotowire and Medical-NER. We select the F1-score of the key and value and structure degree of the output content as metrics. Experimental results show that our method can enable large language models to implement text structuralization task, and can achieve comparable performance.

Our contributions can be summarized as follows:
\begin{enumerate}
    \item We propose an instruction-based text structuralization and information extraction method to make large language models to extract information from texts and output as structured format under zero-shot condition.
    \item For text structuralization and information extraction, we design experiments to verify the effectiveness of our method. We also design experiments to investigate the effect of different instructions on the experimental results.
    \item F1-score and structure degree results of two experiments on Wikibio, WikitableT, Rotowire, Medical-NER, and ACE2005 show that our method can enable large language models to conduct the text structuralization and information extraction task and achieve a comparable performance, without training or manually annotated data.
\end{enumerate}

\section{Related Work}
\textbf{Information Extraction} (IE) is the task of extracting information, which is structured data, from a text, which is unstructured or semi-structured data. There are some traditional sub-tasks in IE. For example, named entity recognition (NER) recognizes entities and corresponding categories appearing in a text. Relation extraction (RE) identifies the relationships between entities and events. Event extraction (EE) distinguishes events described in a text, and is usually divided into Trigger Identification, Argument Identification, Argument Role Classification, etc. Role-filler entity extraction (REE) fills entities into event templates and is similar to EE.

Over the last two decades or more, there has been extensive research into EE and its subtasks. For example, \citep{collins1999unsupervised} introduced a method using the spelling rule that use look-up tables or predefined patterns. \citep{zhou2002named} employed Hidden Markov Model(HMM) to identify the name entity and their types. After these there are more advanced methods like LSTM\citep{jie2019dependency}, and with transformer model \citep{vaswani2017attention} proposed, BERT series models \citep{devlin2018bert} based on transformer encoder achieve the state-of-the-art in NER task, and also perform well in RE and EE tasks(\citealp{wadden2019entity}; \citealp{zhang2019joint}; \citealp{lin2020joint}). Other methods include using dependency trees in deep learning models for RE task (\citealp{miwa2016end}; \citealp{liu2015dependency}; \citealp{zhang2018graph}) and incorporating the interaction between event types \citep{li2019joint} or argument roles \citep{wang2019open} using hierarchy-based modeling for EE task. However, most existing methods or models require a large number of pre-annotated datasets for training, and the generalization ability are not strong.

Another related field is open information extraction (OpenIE) that aims at extracting information from texts without pre-defined schemas ( \citealp{schmitz2012open}; \citealp{stanovsky2018supervised}; \citealp{zhan2020span}). Current OpenIE models can only extract simple structures like tuples from short text. Our method aims at extracting as much information as possible from medium to long texts.

\textbf{Large language models} (LLM) are a series of models that make use of deep learning networks and are trained on a vast quantities corpus(\citealp{brown2020language}; \citealp{wei2022emergent}; \citealp{xu2022systematic}). Large language models have shown impressive performance on various downstream tasks, like machine translation \citep{voita2019bottom}, text generation \citep{yuan2022wordcraft}, and even complex commonsense reasoning \citep{lewkowycz2022solving}. 
GPT-3 \cite{brown2020language}, with 175 billion parameter, has shown strong performance on many NLP tasks and benchmarks in few-shot setting. The latest ChatGPT model released by OpenAI some time ago has set off a new round of AI boom. From technical answers to scene play, from ghostwriting papers to chatting to relieve boredom, ChatGPT seems to be omnipotent. The most common way to use LLMs is fine-tuning, which refers to the processing of adapting a general-purpose model for a specific task or domain, achieved by training LLMs on a smaller dataset relevant to the task \citep{wei2021finetuned}. Recent years, in-context learning makes the use of LLMs more convenient and efficient \citep{min2022rethinking}. Provide LLMs a set of prompts (often input-output pairs), and LLMs can learn the patterns and execute the desired task. However, both fine-tuning and in-context learning require more or less sample data, and  still pose difficulties for the general use of LLMs. So researches of LLMs under zero-shot setting become popular.

\begin{table*}[!t]
\centering
\resizebox{1.8\columnwidth}{!}{
\centering
\begin{tabular}{lp{6cm}p{6cm}}
\Xhline{3\arrayrulewidth}
\textbf{Dataset}            & \textbf{Text} & \textbf{Annotated Answers} \\ \hline
\multirow{2}{*}{WikitableT} & emmanuel jal acted as tulu in a film named " africa united " in 2010.                                                                                                                                                                                                & year: 2010; film: africa united; role: tulu; genre: drama   \\ \cline{2-3}
                            & " i think i like it " was released on 4 august 2010 .           & year: 2010; title: " i think i like it "; release date: 4 august 2010\\ \hline
\multirow{2}{*}{Wikibio}    & william ato ankrah , (born 7th july 1979) better known by his stage name akoo nana is a hiplife musician from ghana . since 2009 , he has been making a large impact in africa 's hiplife scene through his authentic sounding music and captivating stage craft . & Name: William Ato Ankrah; Stage Name: Akoo Nana; Date of Birth: 7th July 1979; Profession: Hiplife Musician; Country of Origin: Ghana; Year of Entry into Music: 2009 \\ \cline{2-3} & matt elliott is an english folk guitarist and singer from bristol , england , who plays dark folk music . he also produced and recorded electronic music under the name the third eye foundation .             & Name: Matt Elliott; Nationality: English; Occupation: Folk Guitarist and Singer; Origin: Bristol, England; Genre: Dark Folk Music; Alias: The Third Eye Foundation \\ \Xhline{3\arrayrulewidth}
\end{tabular}
}
\caption{Four examples are taken from WikitableT and Wikibio respectively. Answers of the examples in WikitableT are the source tabular data in the original dataset. We use string matching and manual inspection to eliminate key-value pairs that have nothing to do with the text. Answers of the examples in Wikibio are from manual annotation.}
\label{dex}
\end{table*}

\textbf{Instruction-tuning} is a training method that fine-tune the model on a collection of NLP datasets expressed via natural language instructions \citep{wei2021finetuned}. This simple method has been shown to improve the zero-shot learning abilities of language models. Follow an instruction, LLMs can perform well on unseen tasks (\citealp{wei2021finetuned}; \citealp{chung2022scaling}; \citealp{xu2022multiinstruct}), and even more close to human needs like writing a poem \citep{chakrabarty2022help}. Our method is inspired by the performance of instruction-tuned LLMs under zero-shot condition.

\section{Methodology}

\subsection{Instruction Method}
We add instructions to the text before it is fed to the large language model. Following the idea of \citep{wei2021finetuned}, we first manually construct a collection containing different forms of instructions. The instruction with the best performance on the validation set of Wikibio is used for the rest experiments in our work. Please refer to the section \ref{analysis} for more details. Specifically, the original input text is converted to $\mathbf{Prefix+Text+Suffix}$. Among them, prefix=\emph{"Please get information form text"} is used to indicate the task, and suffix=\emph{"Output as structured data"} is used to control the output format. And the instruction contents can be transformed according to the task. For example, in non-English text structuralization tasks, instructions are translated into the corresponding language.
\subsection{Language Models} 
We leverage two collections of transformer language models with instruction-tuning. The first is GPT-3, for which we use text-ada-001, text-babbage-001, text-curie-001, and text-davinci-003\footnote{https://openai.com/api/}, which presumably correspond to InstructGPT models with parameters 350M, 1.3B, 6.7B and 175B. We use these models through the OpenAI API. The second is FLAN-T5, for which we use FLAN-T5-Base, FLAN-T5-Large, FLAN-T5-XL and FLAN-T5-XXL, which have 250M, 780M, 3B and 11B parameters. We download FLAN-T5 series models from huggingface\footnote{https://huggingface.co/google/flan-t5-xxl} and use the transformers package to load the model \citep{https://doi.org/10.48550/arxiv.2210.11416}.

\subsection{Dataset Construction} \label{construction}
To the best of our knowledge, there is no text structuralization dataset that extracts all information from texts and outputs it as a key-value format without any pre-defined types or schemas. Therefore, we use automatic script and manual annotation to construct the dataset from table-to-text datasets: (1) Wikibio \citep{lebret2016neural}: gathers 728,321 biographies from English Wikipedia and for each article, it provides the first paragraph and the infobox (both tokenized), (2) WikitableT \citep{chen2020wikitablet}: contains Wikipedia article sections and their corresponding tabular data and various metadata, (3) Rotowire \citep{puduppully2019data}: contains instances composed of a text and two tables from the sports domain, where the text is a report of a basketball game and the two tables represent the scores of teams and players respectively, (4) Medical-NER: contains Chinese electronic medical records describing the status of the patient, from the Xunfei Open Platform\footnote{http://challenge.xfyun.cn/topic/info?type=medical-entity}. We list some examples of automatic and manual annotation in Table \ref{dex} and Appendix \ref{B}.

\paragraph{Automatically labeled dataset}
We automatically construct dataset from WikitableT, in which each sample consists of several key-value pairs and a short sentence. We use the short sentence as source text and the corresponded key-value pairs as target structured data. However, there may be a situation where the value of pair is not included or mentioned in the sentence. So, we check all pairs for each sentence by string matching. We filter out pairs that are not relevant to the sentence and save the rest as answers.

\paragraph{Manually labeled dataset}
If the text length is long, the automatic labeling method above is not applicable, since there is a high probability of missing other key information and many key-value pairs cannot be determined whether related to the text merely by string match. And there are lots of troubles in existing table-to-text datasets. For instance, Wikibio is very dirty and has noise information like \textit{imagesize:150px}. WikitableT only contains structrued data that related to points, rebounds, assists, and other numerical information, but misses match location, match time, next match opponent, etc. So we randomly sample 100 texts from the remaining three datasets above and manually label them. Table \ref{datasets} gives the statistics of these datasets.

\begin{table}[!t]
\resizebox{\linewidth}{!}{
\centering
\begin{tabular}{lccc}
\Xhline{3\arrayrulewidth}
\textbf{Dataset} & \textbf{Num} & \textbf{Avg Tokens} & \textbf{Avg Pairs} \\
\hline
WikitableT  & 2000    & 13.89          & 3.76         \\
Wikibio     & 100     & 101.50         & 7.53         \\
Rotowire    & 100     & 359.00         & 16.40        \\
Medical-NER & 100     & 132.56         & 6.28         \\            
\Xhline{3\arrayrulewidth}
\end{tabular}
}
\caption{Statistics of WikitableT, Wikibio, Rotowire, and Medical-NER datasets, including the number of texts, average tokens and average key-value pairs per text. \label{datasets}}
\end{table}

\section{Experimental Setup}

\subsection{Datasets}
For text structuralization experiment, we make use of datasets constructed in \ref{construction}, which are automatically labeled WikitableT and manually labeled Wikibio, Rotowire, and Medical-NER. For information extraction experiment, 
we choose \textbf{ACE2005} \citep{doddington2004automatic} dataset, which is the most popular dataset among information extraction studies. It has annotations for 599 documents with labels for entities, relations and events. Docuemnts annotated for ACE 2005 are in English, Arabic and Chinese from six different domains, i.e., Newswire, Broadcast News, Broadcast Conversations, Weblog, Usenet News Group, and Conversational Telephone Speech. We choose the English text among them to carry out the experiment. 

\subsection{Baselines}
We compare with the following baseline methods.
\paragraph{Few-shot prompting}
For the text structuralizaton experiment, we consider standard few-shot prompting as a baseline, popularized by \citep{brown2020language}, in which a language model is given in-context exemplars of input-output pairs before outputting a prediction for a test-time example. In this experiment, exemplars are formatted as <\emph{text:} + input, \emph{output:}+output>, and for Wikibio, WikitableT and Rotowire, we select 3-6 samples from the training set, and then manually label the results as a prompt.

\paragraph{Instruction with prompt}
In addition to the few-shot prompting, we also consider combining prompting method and instruction method as a baseline for the text structuralizaton experiment. Specifically, we replace the format of exemplars with <$\mathbf{Prefix}$+input, $\mathbf{Suffix}$+output>. 

\paragraph{Deep learning models}
For the information extraction experiment, we choose the state-of-the-art models or methods on ACE2005 datasets as baselines, including (1) \textbf{PURE} \cite{zhong-chen-2021-frustratingly}, which is essentially built on two independent encoders and merely uses the entity model to construct the input for the relation model, (2) \textbf{PL-Marker} \citep{ye2022packed}, which is a neighborhood-oriented packing strategy that considers the neighbor spans integrally to better model the entity boundary information, and (3) NLI+, which addresses IE tasks as an entailment problem. 

\begin{figure*}[!t]
   \centering
   \includegraphics[width=\linewidth]{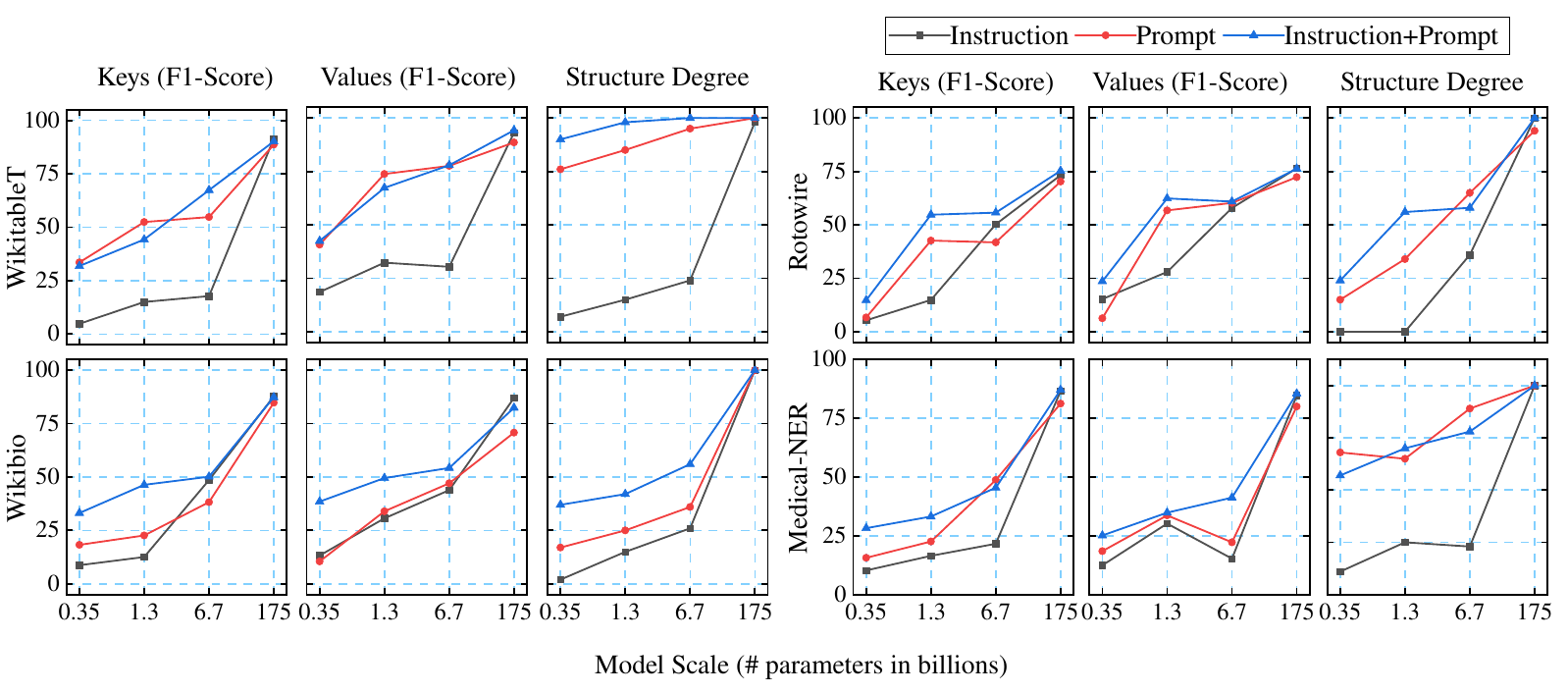}
    \caption{F1-score and structure degree results (\%) of GPT-3 series models using three methods on WikitableT, Wikibio, Rotowire and Medical-NER. Each metric score of three methods generally increase with the expand of the model size.}
    \label{result-gpt}
\end{figure*}

\paragraph{Decoding strategy} Since the maximization-based decoding methods, e.g., greedy search and beam search, tend to generate tedious and repetitive text, we use top-$p$ sampling as our decoding method, following \citet{DBLP:conf/iclr/HoltzmanBDFC20}. We set $P=0.9$.


\subsection{Metrics}
We evaluate the performance of a method based on (1) the number of correct keys, (2) the number of correct values and (3) how structured the output is. For (1) and (2), we adopt the F1 score as the evaluation metric. We set $\mathbf{y}$ to be the set of predicted keys or values, $\mathbf{y}^*$ to be the set of ground-truth. Precision is defined as the percentage of the correctly predicted results among the predicted results: 
\begin{equation}
    P=\frac{1}{|y|} \sum_{y \in \mathbf{y} } \text{max}_{y^* \in \mathbf{y}^*}O(y,y^*)
\end{equation}
Recall is defined as the percentage of the correctly predicted results among the set of ground-truth: 
\begin{equation}
    R =\frac{1}{|\mathbf{y}^*|} \sum_{y^* \in \mathbf{y}^*} \text{max}_{y \in \mathbf{y}} O(y,y^*)
\end{equation}
Finally, $F1 = 2/(1/P+1/R)$. Here, $O(\cdot)$ denotes a way of judging whether the prediction is correct. However, results generated by large language models do not contain types or span boundaries, which are important for evaluation protocol of traditional IE task (\citep{bekoulis2018adversarial}; \citep{taille2020let}). In addition, some metrics used for generation task, such as bleu \citep{papineni2002bleu}, rouge \citep{lin2004rouge}, etc., are not suitable for this situation. For example, both \emph{time} and \emph{date} can describe time information appearing in the text. But the scores of bleu and rouge are 0. So we consider two settings of $O(\cdot)$. For WikitableT, we use the setting in \citep{wu2021text} and consider BERTScore \citep{zhang2019bertscore} to calculate the similarity between predictions and ground-truths, since structured data extracted from WikitableT are relatively fixed and easy. For Wikibio, Rotowire and Medical-NER, we consider human evaluation. For (3), we manually evaluate the structure degree score of the model's generated results for each text, since the delimiters in results vary and are not fixed. The score ranges from 0 to 100, and 100 means that the result is completely composed of data in the form of key-value, and each data is separated by a delimiter. 0 means that the result is a piece of text without any signs of structure.
\begin{figure*}[!t]
   \centering
   \includegraphics[width=\linewidth]{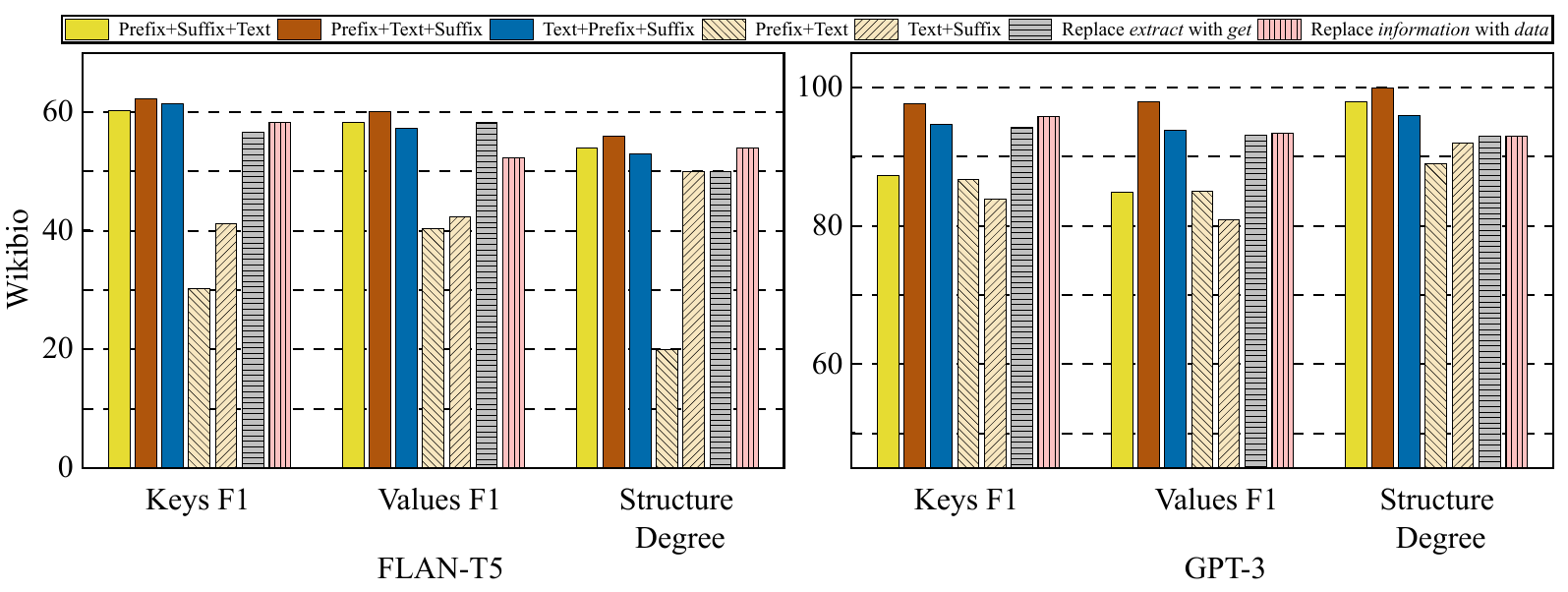}
    \caption{Performance of different instructions for GPT-3 and FLAN-T5 series models on Wikibio.}
    \label{robustness}
\end{figure*}
\section{Results and Analysis}

\subsection{Results of Text Structuralization} \label{analysis}



\paragraph{Main results}~{}
\newline 
Experimental outputs for GPT-3 collection, model size and datasets are shown in Figure \ref{result-gpt}. Due to space limitations, we report the remaining results in the Table \ref{flan-t5-results} in Appendix \ref{appendix_a}. There are three key takeaways. First, as the model scale increases, the F1-Score of the keys and values and the structure degree of the output contents extracted by three methods show an overall upward trend. This suggests that 
a simple instruction can make instruction-tuned large language models perform well in text structuralization task. In addition, the length of the text is also a critical factor. The model performs much better on short texts than long texts. According to our analysis, this may be due to the fact that the excessively long text reduces the proportion of instruction in the input, thus diluting the control effect of instruction on the model's outputs. And FLAN-T5 is just the opposite that owns better performance of instruction method than that of prompting method.

Second, for the GPT-3 series of models, prompting method performs better than instruction method, especially in short to medium texts of Wikibio and WkiTableT. And for a model size of 175B, the results of three methods are very close. For FLAN-T5, the performance of prompting method is lower than that of instruction. The reason is possibly that prompting method is a type of in-context learning,  and GPT-3 develops a wide range of skills and pattern recognition abilities during unsupervised pre-training, and then uses these abilities at inference time to rapidly adapt to or recognize the desired task from prompts.  Therefore, on a low scale, GPT-3 is better at identifying tasks from the prompt than from the instruction. On a larger scale, GPT-3's comprehension ability is powerful enough to recognize tasks from an instruction.

Finally, the results of GPT-3 on Medical-NER show that large language models can extract information from texts in different languages and fields, simply by changing the content of the instruction to the corresponding language.


\begin{table}[!t]
\centering
\resizebox{\linewidth}{!}{
\begin{tabular}{l|l|ccc}
\Xhline{3\arrayrulewidth}
\multicolumn{2}{c|}{\textbf{Models}}  & \textbf{NER} & \textbf{EE} & \textbf{RE} \\\hline
\multicolumn{2}{c|}{PL-MARKER}        & 91.1                 & --                        & 73.0                         \\\hline
\multicolumn{2}{c|}{PURE}             & 90.9                 & --                        & 69.4                         \\\hline
\multicolumn{2}{c|}{NLI+}             & --                   & 71.8                      & --                           \\\hline
FLAN-T5 & Base+I    & 12.70                & 6.37                      & 16.56                        \\
                         & Large+I   & 21.78                & 19.08                     & 20.52                        \\
                         & XL+I    & 29.25                & 29.30                     & 36.99                        \\
                         & XXL+I     & 40.80                & 26.52                     & 33.94                        \\\hline
GPT-3   & text-ada-001+I     & 0                    & 0                         & 0                            \\
                         & text-babbage-001+I & 11.26                & 5.25                      & 0                            \\
                         & text-curie-001+I   & 35.71                & 22.68                     & 15.27                        \\
                         & text-davinci-001+I & 74.31                & 46.95                     & 40.28                \\
\Xhline{3\arrayrulewidth}      
\end{tabular}}
\caption{\label{ace05}
Overall entity, event, and relation F1 scores of baselines and GPT-3 and FLAN-T5 with instruction method (\%) on the test sets of ACE2005. Our method is under zero-shot setting but still achieves comparable performance to the baselines.}
\end{table}

\begin{table*}[!t]

\centering
\resizebox{1.6\columnwidth}{!}{

\begin{tabular}{llcccccc}
\Xhline{3\arrayrulewidth} 
\multirow{2}{*}{\textbf{Setting}} & \multicolumn{1}{c}{\multirow{2}{*}{\textbf{Method}}} & \multicolumn{2}{c}{\textbf{Key F1}} & \multicolumn{2}{c}{\textbf{Value F1}} & \multicolumn{2}{c}{\textbf{Structure Degree}} \\ \cline{3-8}
                                  & \multicolumn{1}{c}{}                                & greedy           & top-p           & greedy            & top-p            & greedy                & top-p               \\ \hline
\multirow{4}{*}{Instruction}      & Base                                                & 24.36            & 34.12            & 33.67             & 43.74             & 33                    & 28                    \\
                                  & Large                                               & 41.17            & 40.4             & 50.27             & 48.84             & 45                    & 43                    \\
                                  & XL                                                  & 40.5             & 46.67            & 66.79             & 73.24             & 60                    & 55                    \\
                                  & XXL                                                 & 45.52            & 49.75            & 71.01             & 76.22             & 61                    & 62                    \\ \hline
\multirow{4}{*}{Prompting}        & Base                                                & 2.36             & 3.67             & 10.37             & 6.25              & 14                    & 2                     \\
                                  & Large                                               & 33.17            & 28.91            & 26.37             & 25.24             & 16                    & 15                    \\
                                  & XL                                                  & 14.28            & 16.39             & 28.26             & 22.04             & 22                    & 12                    \\
                                  & XXL                                                 & 43.30             & 44.78            & 44.23             & 42.86             & 39                    & 36                    \\ \hline
\multirow{4}{*}{Instruction+P}    & Base                                                & 45.26            & 46.34            & 55.26             & 51.89             & 31                    & 35                    \\
                                  & Large                                               & 44.72            & 41.94            & 60.14             & 46.6              & 28                    & 30                    \\
                                  & XL                                                  & 46.38            & 47.87            & 69.81             & 73.47             & 65                    & 62                    \\
                                  & XXL                                                 & 52.57            & 53.14            & 77.23             & 80.45             & 74                    & 73      \\\Xhline{3\arrayrulewidth}              
\end{tabular}
}
\caption{Performance of two decoding strategies used by three methods on Wikibio dataset. The large language model is FLAN-T5 series.\label{decoding}}
\end{table*}
\paragraph{Ablation analysis of instruction-based methods}~{}
\newline
In this subsection, we perform ablation analysis of instruction methods. We consider the impact of the position and content of instruction on the model output. Specifically, we calculate various metrics score of model when the input content is  (1) $\mathbf{Prefix+Suffix+Text}$, (2) $\mathbf{Text+Prefix+Suffix}$, (3) $\mathbf{Prefix +Text}$, and (4) $\mathbf{Text+Suffix}$. In addition, we also replaced keywords in the instruction. We replace \emph{extract} with \emph{get} and \emph{information} with \emph{data} respectively, and also calculate metrics score.

Figure \ref{robustness} shows these results for GPT-3 175B and FLAN-T5-XXL on Wikibio. Changing the position of the prefix and suffix or replacing some words in the instruction causes a small drop in performance. This shows that the large language model is not sensitive to the position of the instruction. It does not rely on position to distinguish instruction from text, nor does it rely on a few specific keywords. It really understands the instruction, recognises it and completes the task. However, with only prefix or suffix, the metric score drops significantly, indicating that both prefix and suffix have an important impact on the performance of the models.

\paragraph{Analysis of decoding strategies}~{}
\newline
Table \ref{decoding} indicates that when the scale of the model is small, the decoding method of greedy decoding is not as good as top-p sampling in F1-score of key and value, but it may have an advantage in structure degree. However, On FLAN-T5-XXL, the performance of the two is very close, showing that the larger the model, the smaller the impact of the decoding method on the output result.

\subsection{Results of Information Extraction}

Table \ref{ace05} shows the results of baselines and GPT-3 series models, from which we can see that for NER, RE, EE tasks, large language models can achieve relatively good performance by adding an instruction. Although there is still a gap compared to baseline models, the exciting thing is that this method does not require any fine-tuning or building of prompts in advance, which reduces the cost. It also shows that large language models have a strong generalisation ability and can perform a variety of different tasks.

\subsection{Case Study}
In order to analyse the shortcomings of instructing a large language model to perform text structuralization task by adding instructions, we collect some bad cases from the outputs of the models and analyse the reasons for their errors. We are pleasantly surprised to find that some errors that are common on small-scale models disappear on larger models. As shown in Table \ref{cases} (Appendix), for case 1, FLAN-T5-XXL extracted the duplicate information \emph{birthdate: 1949} from text, and for case 2, FLAN-T5-XXL misidentified \emph{deathplace} as \emph{zorgvlied}. For case 3, the output information from text-curie-001 is in English, but the text and instructions are all in Chinese. None of these problems occur on GPT-3 175B. This shows that the larger the scale of the model, the fewer mistakes are made and the more accurate the understanding of the instructions. It also shows that the instructions are only a guide and that the results depend entirely on the ability of the model itself and the understanding of the instructions.

\section{Conclusion}
In this paper, we explore the ability of large language models to understand instructions and perform text structuralization task. We propose a method of adding a prefix and a suffix instruction to indicate the desired IE task and structure type, respectively, before feeding the text into a LLM, and we label four datasets manually or automatically. Experiments on datasets from different languages and domains show that a simple instruction can enable large language models to perform comparably to other state-of-the-art methods on datasets, and to generalize to other IE sub-tasks via changing the content of instruction. This method may help researchers build datasets in low-source and domain specific scenarios with low cost, e.g., in the fields of finance and law.


\section*{Limitations}
The limitations of our method mainly come from the disadvantages of using large language models. First of all, most of the large language models that work well are not open source or charged. This makes it difficult for us to conduct batch experiments or daily use on it. Next, a small number of open-source models require a lot of GPU resources when used, which is a difficult problem for quite many researchers, such as students.

\section*{Ethics Statement}
We honor and support the ACL code of Ethics. Text structuralization methods or models aim to extract key information from text and output as structured data. The interaction and assistance process do not involve any bias towards to the participants. All datasets used in this work are from previously published works, and in our view, do not have any attached privacy or ethical issues.

\bibliography{anthology,custom}
\bibliographystyle{acl_natbib}

\appendix
\section{FLAN-T5 Experimental Results}
\label{appendix_a}
This section contains tables for experimental results for FLAN-T5 series models on all datasets for instruction, prompting and instruction with prompt method.

According to our analysis in section \ref{analysis}, we select Top-P sampling as decoding strategy. Since FLAN-T5 series models cannot perform chinese text structuralization task, we do not report the results of FLAN-T5 on Medical-NER dataset. Figure \ref{flan-t5-results} shows the experimental results, from which we can draw different conclusions from the GPT-3 experiment. First, FLAN-T5 owns better instruction learning ability than prompt learning, since for FLAN-T5 series of models, instruction method performs better than prompting method. Second, text structuralization capability appears on any model scale of Flan-T5, and it is strengthened with the increase of the model scale, rather than an emergent ability.

\section{Examples of Rotowire and Medical-NER} \label{B}
\begin{figure*}[!t]
   \centering
   \includegraphics[width=\linewidth]{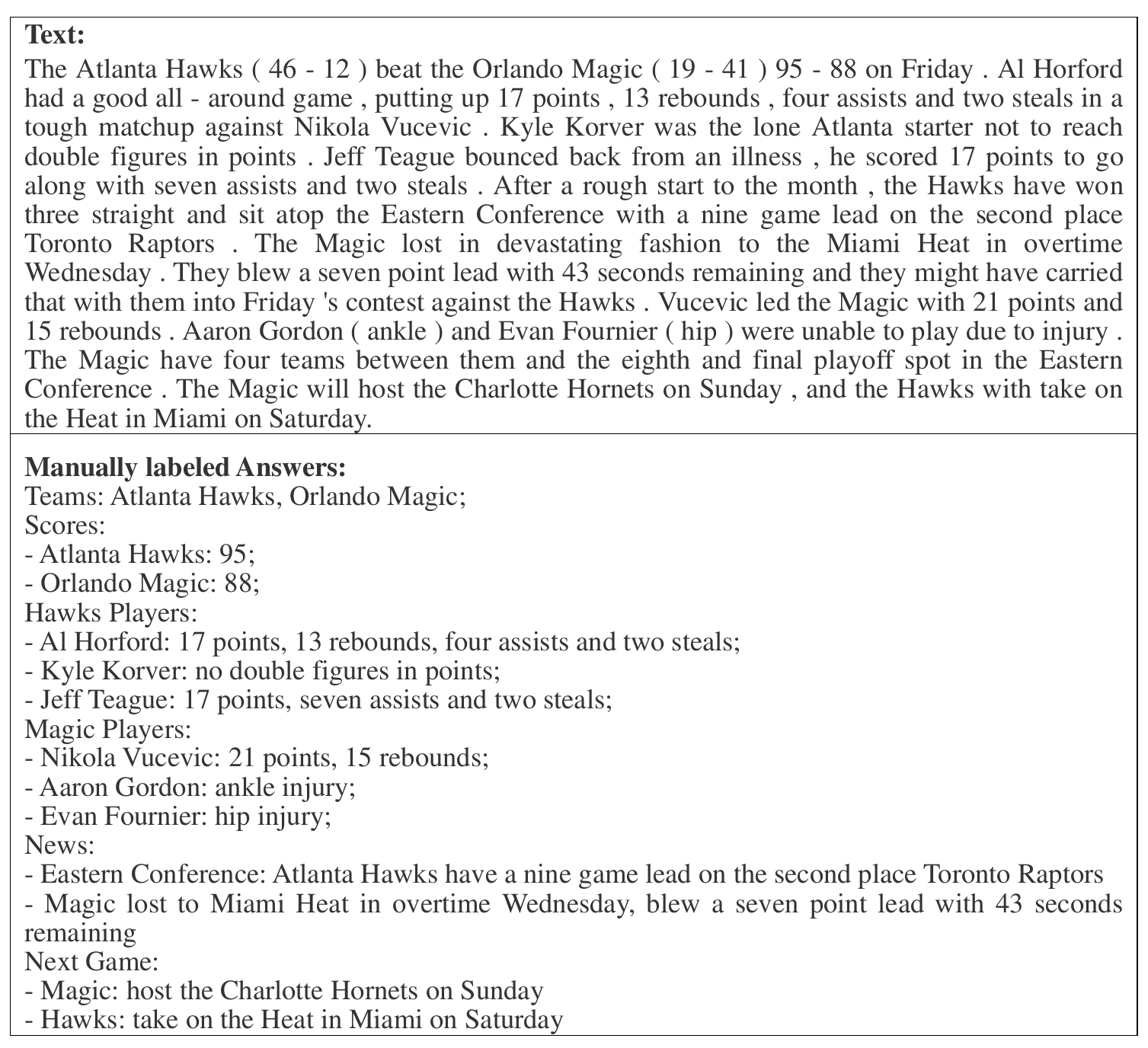}
    \caption{An example of Rotowire.}
    \label{rotowire}
\end{figure*}

\begin{figure*}[!t]
   \centering
   \includegraphics[width=\linewidth]{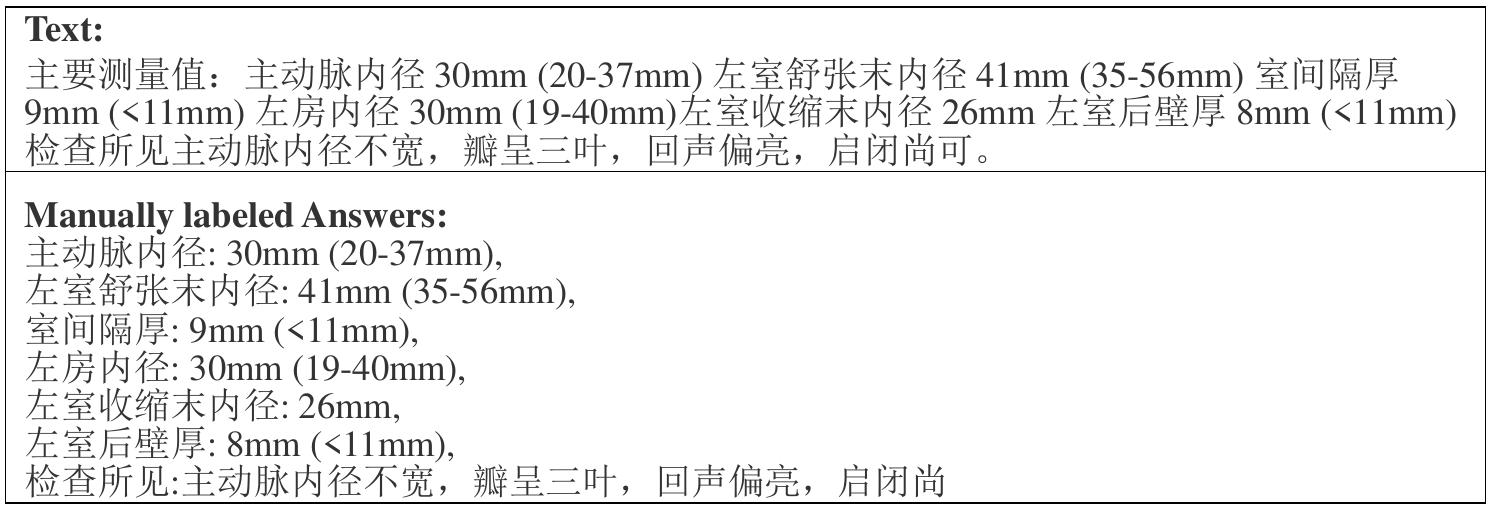}
    \caption{An example of Medical-NER.}
    \label{medical}
\end{figure*}

Figure \ref{rotowire} and \ref{medical} show example of Rotowire and Medical-NER respectively, which are all manually labeled. Since the text of rotowire is too long, we have to use nested structure in labeled answers.
\begin{table*}[!t]
\centering
\resizebox{1.5\columnwidth}{!}{
\begin{tabular}{l|l|cccccc}
\Xhline{3\arrayrulewidth}
\textbf{Size}                & \multicolumn{1}{c}{\textbf{Scale}} & \multicolumn{6}{|c}{\textbf{Datasets}}                                                                                                                                                               \\ \cline{3-8}
                               & \multicolumn{1}{c}{}               & \multicolumn{3}{|c}{Wikibio}                                                                      & \multicolumn{3}{c}{Rotowire}                                                                     \\ \cline{3-8}
& \multicolumn{1}{c}{}               & \multicolumn{1}{|c}{K-F1} & \multicolumn{1}{c}{V-F1} & \multicolumn{1}{c}{S.D.} & \multicolumn{1}{c}{K-F1} & \multicolumn{1}{c}{V-F1} & \multicolumn{1}{c}{S.D.} \\ \hline
\multirow{4}{*}{Instruction}   & Base                               & 37.73                      & 29.2                         & 33                                   & 20.18                      & 13.47                        & 28                                   \\
                               & Large                              & 34.41                      & 29.64                        & 34                                   & 24.46                      & 17.24                        & 43                                   \\
                               & XL                                 & 42.82                      & 37.22                        & 40                                   & 34.51                      & 36.56                        & 55                                   \\
                               & XXL                                & 62.35                      & 60.05                        & 56                                   & 38.22                      & 33.84                        & 62                                   \\ \hline
\multirow{4}{*}{Prompting}     & Base                               & 10.53                      & 14.21                        & 6                                    & 8.23                       & 3.77                         & 7                                    \\
                               & Large                              & 12.85                      & 12.47                        & 15                                   & 4.52                       & 11.24                        & 0                                    \\
                               & XL                                 & 40.76                      & 21.88                        & 28                                   & 9.44                       & 23.58                        & 17                                   \\
                               & XXL                                & 36.29                      & 28.72                        & 25                                   & 25.52                      & 37.78                        & 35                                   \\ \hline
\multirow{4}{*}{Instruction+P} & Base                               & 45.86                      & 35.12                        & 36                                   & 17.35                      & 20.26                        & 29                                   \\
                               & Large                              & 36.92                      & 27.14                        & 40                                   & 19.47                      & 14.17                        & 20                                   \\
                               & XL                                 & 40.33                      & 35.66                        & 54                                   & 15.26                      & 18.54                        & 42                                   \\
                               & XXL                                & 60.13                      & 57.04                        & 70                                   & 33.28                      & 32.75                        & 67                               \\\Xhline{3\arrayrulewidth}   
\end{tabular}
}
\caption{\label{flan-t5-results}
Performance of three methods for FLAN-T5 series on Wikibio and Rotowire. We make use of Top-P sampling decoding strategy.}

\end{table*}

\begin{table*}[!t]
\resizebox{\linewidth}{!}{
\centering
\begin{tabular}{lp{4cm}|l|p{8cm}}
\Xhline{3\arrayrulewidth}      
\textbf{Number}         & \textbf{Text}  & \textbf{Models} & \textbf{Outputs}   \\ \hline
\multirow{2}{*}{Case 1} & \multirow{2}{4cm}{ leonard shenoff randle ( born february 12 , 1949 ) is a former major league baseball player . he was the first-round pick of the washington senators in the secondary phase of the june 1970 major league baseball draft , tenth overall .}  & FLAN-T5-XXL           & leonard randle, birthYear is 1949; leonard randle, birthDate is 12 february 1949 \\ \cline{3-4} &   & GPT-3 text-davinci-003           & \begin{tabular}[c]{@{}l@{}}Name: Leonard Shenoff Randle \\ DOB: February 12, 1949 \\ Professional Baseball Player: Yes \\ Team: Washington Senators \\ Draft Round: First Round \\ Draft Year: 1970\end{tabular}  \\\cline{1-4}
\multirow{2}{*}{Case 2} & \multirow{2}{4cm}{jacoba surie ( 5 september 1879 -- 5 february 1970 ) was a dutch painter . surie was born in amsterdam and trained at the rijksakademie van beeldende kunsten there , where she studied under coba ritsema and joseph mendes da costa . ... surie died in amsterdam and was buried in zorgvlied .} & FLAN-T5-XXL     & \begin{tabular}[c]{@{}l@{}}jacoba surie, deathPlace zorgvlied \\  deathPlace, amsterdam\\  birthPlace, amsterdam\\  deathDate, 5 february 1970\\  birthDate, 5 september 1879\end{tabular} \\\cline{3-4}
& & GPT-3 text-davinci-003  & \begin{tabular}[c]{@{}l@{}}Name: Jacoba Surie\\ Date of Birth: 5 September 1879\\ Date of Death: 5 February 1970\\ Place of Birth: Amsterdam\\ Place of Death: Amsterdam\\ Training: Rijksakademie van Beeldende Kunsten\\ Mentors: Coba Ritsema, Joseph Mendes Da Costa\\ Membership: Arti et Amicitiae, Pulchri Studio\\ Notable: Amsterdamse Joffers\\ Burial: Zorgvlied\end{tabular} \\\hline
\multirow{2}{*}{Case 3} & \multirow{2}{4cm}{\begin{CJK*}{UTF8}{gbsn}99mTc-MIBI双时相法：15分钟早期像示甲状腺显影较淡，位置、形态正常，甲状腺右叶下极部位见异常放射性浓聚，余部未见明显异常放射性分布区；60、240分钟延迟像示甲状腺右叶下极部位仍有放射性浓聚,...,考虑为甲状旁腺腺瘤显影可能。\end{CJK*}} & GPT-3 text-curie-001           & The size and shape of the right hepatic lobe is normal. There is no enlargement or distortion of the liver silhouette. The right lobe has many small, low-density focal areas with few if any enhancement features. The density of the focal areas is significantly increased over time, and the edge of the focal area becomes more robust. \\\cline{3-4}
&    & GPT-3 text-davinci-003  & \begin{tabular}[c]{@{}l@{}}\begin{CJK*}{UTF8}{gbsn}甲状腺显影： \end{CJK*}\\ \begin{CJK*}{UTF8}{gbsn} - 99mTcO4- ：正常，未见明显异常放射性分布区 \end{CJK*}\\ \begin{CJK*}{UTF8}{gbsn} - 99mTc-MIBI双时相法：\end{CJK*}\\ \begin{CJK*}{UTF8}{gbsn}   - 15分钟早期：较淡，位置、形态正常，右叶下\end{CJK*}\\\begin{CJK*}{UTF8}{gbsn}极部位见异常放射性浓聚，余部未见明显异常放\end{CJK*}\\\begin{CJK*}{UTF8}{gbsn}射性分布区 \end{CJK*}\\  \begin{CJK*}{UTF8}{gbsn}  - 60分钟延迟：右叶下极部位仍有放射性浓聚，\end{CJK*}\\\begin{CJK*}{UTF8}{gbsn}余部甲状腺影消退 \end{CJK*}\\ \begin{CJK*}{UTF8}{gbsn}诊断建议：考虑为甲状旁腺腺瘤显影可能\end{CJK*}\end{tabular} \\ \Xhline{3\arrayrulewidth}                                                                                        
\end{tabular}
}
\caption{\label{cases}Cases from outputs of different models.}
\end{table*}

\end{document}